\documentclass[11pt]{article}

\usepackage{acl}

\usepackage{latexsym}
\usepackage[T1]{fontenc}
\usepackage{tgheros}
\usepackage{tgcursor}

\usepackage[T1]{fontenc}

\usepackage[utf8]{inputenc}

\usepackage{microtype}
\usepackage{amsfonts}
\usepackage{url}
\usepackage{dirtytalk}
\usepackage{float}
\usepackage{hyperref}
\usepackage{enumitem}
\usepackage{color}
\usepackage{inconsolata}
\usepackage{todonotes}
\usepackage[normalem]{ulem}
\usepackage{booktabs}
\usepackage{subcaption}
\usepackage{hyperref}
\usepackage{url}
\usepackage{graphicx}
\usepackage{multirow}
\title{Multilingual and Multimodal LLMs in the Wild: Building for Low-Resource Languages}

\author{Firoj Alam, Shammur Absar Chowdhury, Enamul Hoque Prince\\
  Qatar Computing Research Institute, HBKU, Qatar, York University, Canada \\
  \texttt{\{fialam, shchowdhury\}@hbku.edu.qa}, enamulh@yorku.ca \\
  \url{https://mm-llms-in-the-wild.github.io}
}

\begin{document}
\maketitle
\section{Abstract}

Multimodal LLMs are evolving from vision–language to tri-modality that \emph{see, hear, and read}, yet pipelines and benchmarks remain English-centric and compute-heavy. The tutorial offers an overview of this emerging research area for \emph{multilingual} \emph{multimodality} across \textbf{text, speech, and vision} under limited data/compute budgets, synthesizing foundations,
recent multilingual models (PALO, Maya), speech–text LLMs.
We cover low-cost data creation/curation; adapter stacks 
for tri-modal alignment; \emph{culture-aware} evaluation beyond English 
and hands on resources for fine-tuning a compact multilingual VLM and wiring a speech$\rightarrow$text$\rightarrow$LLM pipeline. 
The content will be delivered as an interactive half-day tutorial, designed for researchers and practitioners working on multilingual, multimodal AI in low-resource language settings. 

\section{Introduction}

Multimodal large language models (LMMs) are transforming how we build AI systems: instead of only reading text, they can now \emph{see}, \emph{hear}, and \emph{read}—integrating visual, audio, and textual signals under a unified architecture \cite{alam2025everydaymmqa}. Surveys show that while LMMs are advancing rapidly, most datasets, benchmarks, and models remain heavily English-centric and optimized for high-resource languages and settings \cite{yin2024survey,wu2023survey}. This leaves many low-resource languages, dialects, and culturally-specific modalities under-served, especially in multilingual, speech-and-vision contexts.

The multilingual and multi-modal AI community -- spanning NLP, speech processing, and computer vision, stands at a critical juncture. On one hand, we have foundation models (e.g., vision–language systems) that demonstrate strong capabilities in high-resource contexts; on the other, communities working with under-represented languages lack tools, datasets, and evaluation protocols that support tri-modal workflows (text, speech, vision) in resource-constrained environments. Recent work (e.g., PaLI, mSLAM) shows promise for multilingual cross-modal pre-training \cite{chen2022pali,bapna2022mslam}, but large gaps remain in end-to-end pipelines, culturally grounded benchmarks, and efficient architectures tailored for real-world low-resource deployment.

In this tutorial we aim to bring together the state-of-the-art in multilingual multimodal LMMs with practical, reproducible recipes for data creation (covering, text, speech, vision from different disciplines), model alignment, fine-tuning, and evaluation, focusing explicitly on low-resource languages and cultural contexts. We will cover foundational models and architectures, examine evaluation and bias considerations, and provide hands-on labs for fine-tuning compact multilingual vision–language models and wiring speech → text → LLM pipelines. Our goal is to equip researchers and engineers working in multilingual, multi-modal AI with actionable tools, checklists, and benchmarks to deliver inclusive, grounded AI systems beyond the high-resource norm.

\subsection{Foundations of Multimodal and Multilingual Models}
We review the evolution of large models: from text-only LLMs based on the Transformer architecture \cite{vaswani2017attention}, to vision-language models (e.g., BLIP-2, LLaVA) and now to unified speech–text–vision systems (e.g., PaLM-E, AudioPaLM). We highlight how multilingual extensions of these, while nascent, open pathways for inclusive AI in under-represented languages. We also discuss how modalities add complementary information, e.g., visual or audio context can compensate for scant textual data in low-resource languages \cite{lupascu2025large}.

\subsection{Challenges in Low-Resource Multimodality}
Lower-resource languages bring numerous obstacles: scarcity of paired text/image/audio data, missing benchmarks, disparate dialects/scripts, and compute constraints for large models. For instance, a survey found that out of 106 studies in low-resource multimodal learning across 75 languages, the vast majority focused on text+image pairs but neglected audio or video modalities \cite{lupascu2025large}. We analyse key bottlenecks, data creation, modality alignment, adaptation of pre-trained models, and evaluation frameworks, and how these hinder deployment in real-world multilingual/multimodal settings.

\subsection{Dialectal Challenge}
Collecting dialectal multimodal data at scale required tackling issues that standard VQA corpora rarely face. In \emph{EverydayMMQA}/OASIS, we had to \textit{(i)} recruit and balance native speakers across 18 countries and Arabic varieties; \textit{(ii)} capture \emph{spoken} questions that naturally include dialectal phonology, code-switching, and region-specific lexicon; \textit{(iii)} align speech/text prompts with culturally grounded images beyond object labels; and \textit{(iv)} verify answers under divergent orthographies and local conventions. On the evaluation side, our benchmarks emphasize pragmatic, commonsense, and culturally aware aspects and, across four input modes (speech-only, text-only, speech+image, text+image), surface failure modes typical of dialectal settings (e.g., misinterpretation of dialectal terms, sensitivity to ASR noise, weaker transfer from English). Overall, the study highlights significant gaps between current general-purpose multimodal LLMs and the demands of dialect-rich, everyday queries, and motivates training and evaluation pipelines explicitly grounded in local culture.

\subsection{Relevance for Our Community}
This tutorial is timely for researchers and practitioners working at the intersection of multilingual NLP, speech processing, and vision. Its relevance spans:
\begin{itemize}[noitemsep,topsep=0pt,leftmargin=*,labelsep=.5em] 
  \item Building multimodal datasets and models for under-served languages (e.g., Arabic dialects, Indic, African, Southeast Asian languages).
  \item Adapting large vision–language–speech models with limited data and compute via techniques like PEFT, adapters, and MoE.
  \item Evaluating models beyond English: culture-aware benchmarks, dialect resilience, and modality robustness.
  \item Deploying inclusive and accessible systems, with awareness of language diversity, script/dialect variation, and societal context.
\end{itemize}


\subsection{What this tutorial offers}  
This tutorial guides participants through building inclusive, multilingual, multimodal systems (text, speech \& vision) for low-resource settings. We begin with the evolution of LLMs, covering their architecture, multilingual extensions, and multimodal variants. Then we explore concrete model families: vision–language (e.g., PALO), speech–text LMMs (SeamlessM4T, AudioPaLM), and multilingual multimodal models (e.g., Maya). Next, we dive into hands-on methods: low-cost data creation and curation (translation, weak supervision, filtering); efficient training (PEFT, adapters, Mixture-of-Experts); and tri-modal alignment workflows. We then examine evaluation and deployment: multilingual cultural benchmarks (xGQA, HaVQA), visualization datasets (FigureQA, CharXiv), dialectal tests, and addressing hallucination, bias, toxicity, as well as compute/latency trade-offs. An outline of the tutorial is reported in Section \ref{sec:outline}.

\section{Type of the Tutorial}
The tutorial is both introductory, covering a number of topics related to the capabilities of LLMs, but it is also cutting-edge, covering some latest developments in these areas. 
Attendees will have an overview of tasks, languages, dialects and modalities related to LLMs, which will put them up to speed to do research in the area. The tutorial targets anyone interested in employing LLMs for NLP, speech and multimodal tasks. We believe researchers working on lower-resource languages will be especially interested. 

\section{Tutorial organisers}
\begin{itemize}[noitemsep,topsep=0pt,leftmargin=1.0em,labelsep=.5em]
\item \href{http://sites.google.com/site/firojalam/}{\bf Firoj Alam} is a Scientist at the Qatar Computing Research Institute (QCRI), HBKU.
\item \href{http://shammur.one/}{\bf Shammur Absar Chowdhury} is a Scientist at QCRI, HBKU.
\item \href{https://www.yorku.ca/enamulh/}{\bf Enamul Hoque} is an Associate Professor at York University.
\end{itemize}

\section{Target Audience}
Researchers and practitioners in NLP, speech, and vision, especially those building for low-resource languages, dialects, or culturally grounded domains. Suitable for graduate students, academic/industry ML engineers, and dataset curators with basic LLM fine-tuning experience. Minimal prior speech or vision background is required; we provide extensive materials. Attendees seeking practical PEFT recipes 
multimodal adapters, and robust, culture-aware evaluation will benefit most.

\section{Outline}
\label{sec:outline}

Below, we offer an outline of the tutorial. More information and materials will be available online on the tutorial website upon acceptance.

\subsection{Introduction}
\begin{enumerate}[label=(\roman*), itemsep=0pt, topsep=0pt]
    \item Why tri-modality (text--vision--speech) matters for low-resource \& Global South contexts
    \item From vision--language to general-purpose multimodality: BLIP-2, LLaVA, KOSMOS-1, PaLM-E
    \item What this tutorial adds: multilingual focus, speech integration, efficiency (PEFT/MoE), culture-aware evaluation
\end{enumerate}
\noindent \emph{\textbf{References}:} \cite{li2023blip2,liu2023visual,kosmos1,dries2023palme}

\subsection{Multilingual and Multimodal Models}
The following are representative models; we will cover additional systems in the tutorial.
\begin{enumerate}[label=(\roman*), itemsep=0pt, topsep=0pt]
    \item Vision--language (multilingual)
        \begin{enumerate}[label=(\alph*), itemsep=0pt, topsep=0pt]
           \item PALO (10-language LMM), Maya (8-language, toxicity-aware data pipeline)
        \end{enumerate}
    \item Speech--text LMMs
        \begin{enumerate}[label=(\alph*), itemsep=0pt, topsep=0pt]
           \item SeamlessM4T (unified S2S/S2T/T2S/T2T/ASR), AudioPaLM (joint speech+text)
        \end{enumerate}
    \item Robust ASR backbones for pipelines
        \begin{enumerate}[label=(\alph*), itemsep=0pt, topsep=0pt]
           \item Whisper (weakly supervised, multilingual)
        \end{enumerate}
\end{enumerate}
\noindent \emph{\textbf{References}:} \cite{maaz2024palo,alam2024maya,seamlessm4t,rubenstein2023audiopalm,radford2022robust}

\subsection{Multilingual \& Multimodal Resource Development}
\subsubsection{Multilingual \& Multimodal Resources}
\begin{enumerate}[label=(\roman*), itemsep=0pt, topsep=0pt]
    \item Low-cost pipelines: translation \& back-translation, weak supervision, OCR/ASR bootstraps
    \item Safety \& culture: toxicity filtering, demographic balance, licensing/consent (Maya case study)
    \item Resource orientation: multilingual V+L evaluation sets (xGQA, MaRVL, HaVQA) and training data considerations
\end{enumerate}
\noindent \emph{\textbf{References}:} \cite{alam2025everydaymmqa,alam2024maya,pfeiffer2022xgqa,liu2021marvl,parida2023havqa}

\subsubsection{Reasoning Across Visual and Structured Modalities}
\begin{enumerate}[label=(\roman*), itemsep=0pt, topsep=0pt]
    \item Visualization datasets (FigureQA, CharXiv, ChartQAPro, DashboardQA))
    \item Reasoning Techniques: multimodal chain-of-thought, ReAct prompting, and structured decoding for spatial/tabular data.
\end{enumerate}
\noindent \emph{\textbf{References}:} 
\cite{kahou2018figureqa,wang2024charxiv,masry2025chartqapro,kartha2026dashboardqa}

\subsection{Architectures and Efficient Training}
\begin{enumerate}[label=(\roman*), itemsep=0pt, topsep=0pt]
    \item Adapter/projector stacks for VLMs (e.g., BLIP-2 Q-Former), early vs.\ late fusion
    \item PEFT in practice (LoRA/QLoRA), quantization notes for constrained VRAM
    \item Mixture-of-Experts for modality/language specialization: MoME, Uni-MoE; routing \& capacity factors
\end{enumerate}
\noindent \emph{\textbf{References}:} \cite{li2023blip2,shen2024mome,li2024unimoe}


\subsection{Speech-centric LLMs}
\begin{enumerate}[label=(\roman*), itemsep=0pt, topsep=0pt]
    \item Wiring speech$\rightarrow$text$\rightarrow$LLM for VQA/QA; streaming, VAD/diarization hooks
    \item Unified speech--text LMMs vs.\ cascades: deployment trade-offs (latency, robustness, coverage)
\end{enumerate}
\noindent \emph{\textbf{References}:} \cite{radford2022robust,seamlessm4t,rubenstein2023audiopalm}

\subsection{Evaluation, 
Benchmarking Resources, 
Error analysis
}
\begin{enumerate}[label=(\roman*), itemsep=0pt, topsep=0pt]
    \item Culture-aware, multilingual benchmarks and diagnostics (xGQA, MaRVL, HaVQA)
    \item Stress tests: dialect shifts, noise/occlusion, OCR-heavy inputs, hallucination \& grounding checks
\end{enumerate}
\noindent \emph{\textbf{References}:} \cite{pfeiffer2022xgqa,liu2021marvl,parida2023havqa}

\subsection{Resources and demo applications \& wrap-up [25 min]}
\begin{enumerate}[label=(\roman*), itemsep=0pt, topsep=0pt]
    \item LoRA-tune a compact multilingual VLM; quick eval on a culture-aware slice
    \item Speech front-end (Whisper/Seamless) into an instruction-tuned LLM; measure ASR$\rightarrow$task impact
\end{enumerate}
\noindent \emph{\textbf{References}:} \cite{liu2023visual,seamlessm4t}

\section{Technical Requirements}
No special requirements; we will use standard A/V setups provided by the organizers.

\section{Diversity Considerations} 
We will run an inclusive tutorial that reflects linguistic, cultural, geographic, and disciplinary diversity.


\noindent
\textbf{Contribution to academic diversity.}
The tutorial centers multilingual, multimodal \emph{everyday knowledge} and explicitly links language, speech, and vision communities. We foreground low-resource and culturally specific contexts and encourage collaboration across academia, industry, and public-interest groups.

\noindent
\textbf{Representation.}
We will actively advertise the tutorial globally, with special outreach to under-represented regions and communities, to ensure participation from all corners of the world. Our outreach will include different mailing lists, and social media.


\section{Reading List}
Relevant papers has been listed in Section \ref{sec:outline}.

\section{Presenters} 

\paragraph{\href{http://sites.google.com/site/firojalam/publications}{Firoj Alam}} (Qatar Computing Research Institute, HBKU) is a Senior Scientist and a senior IEEE and ACM member. 
He has been co-organizing several workshops and shared tasks, including BLP-2023 and GenAI Content Detection at COLING-2025. His shared task experience includes the CheckThat! Lab at CLEF (2021-2025), NLP4IF 2021, and SemEval 2021 Task 6. He has co-oranized a tutorial at EACL-2024. He has also served as a PC and SPC member for numerous conferences ($^*$ACL, AAAI, IJCAI, ICWSM, etc.), as well as a publications and panel chair and has volunteered as a reviewer for various journals. 


\paragraph {\href{http://shammur.one/}{Shammur Absar Chowdhury}} (Qatar Computing Research Institute, HBKU) is a scientist. She broadly works in the area of Conversational AI encompassing language acquisition, speech discourse, code-switching speech recognition and turn-taking in a spontaneous human conversation. 
Shammur has been contributing to strengthen the community by organizing workshops and shared tasks like IEEE SLT Workshop 2022 (local chair), JSALT Summer workshop 2022 (Multilingual and CS ASR), SemEval-2022 (Task3) among others. She has serve as PC and SPC member of many top NLP and Speech conferences -- ACL, Interspeech, ICASSP, NAACL, EMNLP etc; and volunteered as a reviewer for various international journals.

\paragraph {\href{https://www.yorku.ca/enamulh}{Enamul Hoque Prince}} (York University) is an Associate Professor, and Director of the Intelligent Visualization Lab. His research brings together information visualization, natural language processing, and human–computer interaction to make data exploration more accessible, inclusive, and responsible. His recent work focuses on multimodal large language models and agentic frameworks for data visualization and analytics, leading to widely used benchmarks and models. He serves the community as an Area Chair for the ACL Rolling Review and on the IEEE VIS program committee.



\section{Other Information}
At EACL 2024, we organized a closely related tutorial \cite{alam2024llms} that drew 354 registrants, with 40+ live attendees (in person and online). We have also run Birds-of-a-Feather sessions at EMNLP and COLING on similar themes, attracting 30–50 participants per session. Accordingly, we expect 40+ attendees for this edition.

\section{Ethics Statement}
Our tutorial is based on our own work in the area, related studies and public sources. Credit will be given wherever needed. Any biases are unintended.


\bibliography{bibliography}

\end{document}